\documentclass[twoside,11pt,preprint]{article}
\usepackage{blindtext}
\usepackage{dmlr2e}
\usepackage[frozencache,cachedir=minted-cache]{minted}

\usepackage{graphicx} 

\usepackage[utf8]{inputenc} 
\usepackage[T1]{fontenc}    
\usepackage{hyperref}       
\usepackage{url}            
\usepackage{booktabs}       
\usepackage{amsfonts}       
\usepackage{nicefrac}       
\usepackage{microtype}      
\usepackage{xcolor}         
\usepackage{minted} 

\usepackage{multirow}
\usepackage{subcaption}
\usepackage{tablefootnote}
\usepackage{algorithm}
\usepackage{algpseudocode}

\usepackage{tikzexternal}

\usepackage{pifont}
\newcommand*\colourcheck[1]{%
  \expandafter\newcommand\csname #1check\endcsname{\textcolor{#1}{\ding{52}}}%
}

\definecolor{ForestGreen}{RGB}{34, 139, 34}
\colourcheck{ForestGreen}
\definecolor{Red}{RGB}{255, 0, 0}
\expandafter\newcommand\csname Redcross\endcsname{\textcolor{Red}{\ding{55}}}
\newcommand{\tocite}[1]{\textcolor{red}{[MISSING CITATION]}}
\newcommand{\citen}[1]{[\citenum{#1}]}

\usepackage{lastpage}
\dmlrheading{23}{2024}{1-\pageref{LastPage}}{9/23; Revised 10/23}{11/23}{21-0000}{Bryce Ferenczi, Rhys Newbury, Michael Burke, Tom Drummond} 
\ShortHeadings{Carefully Structured Compression: Efficiently Managing StarCraft II Data}{Ferenczi, Newbury, Burke and Drummond}

\firstpageno{1}

\begin{document}

\title{Carefully Structured Compression: Efficiently Managing StarCraft II Data}

\author{%
  Bryce Ferenczi \email bryce.ferenczi@monash.edu\\
  \addr Monash University, Melbourne, Australia\\
  \AND
  Rhys Newbury \email rhys.newbury@monash.edu\\
  \addr Monash University, Melbourne, Australia\\
  \AND
  Michael Burke \email michael.g.burke@monash.edu\\
  \addr Monash University, Melbourne, Australia\\
  \AND
  Tom Drummond \email tom.drummond@unimelb.edu.au\\
  \addr University of Melbourne, Melbourne, Australia\\
}

\editor{My editor}

\maketitle

\begin{abstract}%
Creation and storage of datasets are often overlooked input costs in machine learning, as many datasets are simple image label pairs or plain text. However, datasets with more complex structures, such as those from the real time strategy game StarCraft II, require more deliberate thought and strategy to reduce cost of ownership. We introduce a serialization framework for StarCraft II that reduces the cost of dataset creation and storage, as well as improving usage ergonomics. We benchmark against the most comparable existing dataset from \textit{AlphaStar-Unplugged} and highlight the benefit of our framework in terms of both the cost of creation and storage.
We use our dataset to train deep learning models that exceed the performance of comparable models trained on other datasets. The dataset conversion and usage framework introduced is open source\footnote{\url{https://github.com/5had3z/sc2-serializer}} and can be used as a framework for datasets with similar characteristics such as digital twin simulations. Pre-converted StarCraft II tournament data is also available online.\footnote{\url{https://bridges.monash.edu/articles/dataset/Tournament_Starcraft_II/25865566}}
\end{abstract}


\section{Introduction}

Deep learning algorithms rely on two fundamental components: model architecture and data. The volume of data required for a model to effectively capture the underlying patterns therein can be substantial~\citen{NEURIPS2022_c1e2faff}. This often serves as a limiting factor for model performance. Data introduces associated costs across different stages: collection, storage and usage. Collection costs can include non-trivial acquisition and annotation. Storage costs are primarily dictated by the overall size of the dataset. Data usage encompasses the computational resources needed to pre-process data for learning and the engineering effort required to implement necessary transformations.

For tasks with a relatively simple structure, such as predicting an enumerated class from an image, the dataset structure itself is a straightforward mapping between input data and labels, which could be represented by the subfolder containing the target image~\citen{ImageNET}. Tasks involving more complex structures, such as physics simulations~\citen{JetClass}, require deliberate handling to prevent increased storage and access costs that naive algorithms would incur, such as uncompressed padding of variably sized arrays to create a regular shape. This complexity in data structure is also apparent in video games with time-varying state space dimensions. Video games have served as a popular basis for developing a variety of deep-learning-based algorithms, ranging from outcome prediction~\citen{SC2EGSet, Vinyals2019, sanchez-predict-games, kulinski2023starcraftimage}, detecting cheaters~\citen{fpsCheating}, identifying toxicity~\citen{toxicGameplay} and training agents that play them~\citen{mathieu2023alphastar}. Video game entertainment is a massive industry projected to reach a value of US$\$282.30$bn~\citen{gameTAM}, and it is beginning to leverage advancements in generative artificial intelligence~\citen{cao2023texfusion,videoGameLLM} and reinforcement learning (RL)~\citen{mathieu2023alphastar,battlefieldRL}. These developments are not only enhancing player experiences, but also contributing to research areas outside of gaming such as digital twin simulations.

Applying machine learning to a complex task space often necessitates a substantial amount of data. Running a live simulation enables generation of this data on-the-fly, however it can be more compute intensive than loading pre-recorded data. When using pre-recorded data, the cost of storing a sufficient amount of data for training can become a limitation. StarCraft II (SC2) exemplifies this live simulation versus recording problem. SC2 records games in relatively small \textit{replay files}, containing sequences of player actions and related metadata, but lacks state information necessary for model training. Acquiring state information requires replaying the game simulation, a computationally expensive task that can be done offline. However, recorded state data inflates the dataset size orders of magnitude greater than the original list of actions. We identify key considerations when structuring data for serialization to improve compressibility of SC2 data and make the following contributions:

\begin{itemize}
    \item We reduce the size of SC2 data by $\approx90\%$ compared to \citet{mathieu2023alphastar}.
    \item We open source a SC2 replay serialization tool with reduced memory and compute overhead compared to \citet{mathieu2023alphastar}; doubling serialization throughput for systems with $\leq4$GB RAM per CPU core.
    \item Game outcome predictors trained with this dataset exceed previous methods in accuracy by an absolute $\approx11\%$ (\cite{vinyals2017starcraft} versus CNN+MLP).
    \item A novel in-game minimap forecasting model achieves an Area-Under-the-Curve (AUC) of $0.923$ when forecasting $9$ seconds into the future.
\end{itemize}

Though we primarily focus on evaluating on SC2 data, these findings are transferable to scenarios with a similar structure. Large scale digital twin scenarios, such as in smart manufacturing or supply chain, also consist of a temporally dynamic set of data nodes, such as sensors, robots and human workers, with various attributes and complex interactions with one-another. \citet{drissi2023role} note lack of research in modeling intralogistics processes and facility scale applications of machine learning. The findings of this paper and included data framework can improve accessibility in applying deep learning techniques on large, complicated digital twin datasets. The open-source code has been designed to be repurposed in new scenarios with minimal changes.

\section{Related Work}

\subsection{Preliminaries} \label{sec:prelim}

StarCraft II is a multiplayer real-time strategy game produced by Blizzard and has a highly competitive esports league with hundreds of thousands of dollars in prize money at stake \citen{katowice-prize}. Players collect resources to build and expand their base and army and defeat their opponents. There are three playable factions (Zerg, Terran and Protoss), each with their own unique play style. Although there are builtin AI players that use a rules-based system for control, these are unable to compete against humans and consequently rely on in-game advantages at higher difficulties. Due to the game's observation- and action-space complexity, StarCraft II has been a popular environment for developing machine learning based AI\footnote{Querying academic search engines with ``StarCraft 2'' publications since '20 and returned 275 results from ScienceDirect, 13,100 from GoogleScholar, 51 from Web of Science and 231 from arXiv.org (17th Aug '24).}. \textit{AlphaStar}~\citen{Vinyals2019} is the most renowned attempt, achieving `grandmaster' status. \textit{AlphaStar} is bootstrapped, before a self-play phase against other agents, through behavior cloning: predicting the action of a player conditioned on the cumulative state of the game, with an optional latent variable that represents the latent strategy of the human. Further work, \textit{AlphaStar-Unplugged} \citen{mathieu2023alphastar}, solely uses offline RL to train a model and reliably beats the previous behavior cloning model with a win rate of 90\%. 

Lossless compression algorithms need to reconstruct the original data faithfully. ZLIB~\citen{zlib} uses a combination of LZ77~\citen{ziv1977universal} and Huffman coding~\citen{huffman1952method}. LZ77 finds repeated structures in data to replace with a reference to a previous instance, effectively reducing the overall size. Huffman coding replaces the ``words'' of the data with a different representation, depending on its frequency (fewer bits for more common ``words''). Data is more compressible when there are more frequent and repeated structures and measured by the compression ratio between the original and compressed size. The compression ratio yield from data with many repeated patterns will be greater than that of data with fewer patterns. The compression ratio of a given dataset can be improved by prepossessing it into a structure that complements the compression algorithm.

Two of the fundamental data types in computing are a structure (\textit{struct}) and an array. A \textit{struct} is a collection of elements that that may represent a single object. For example, a unit in a video game can be described by a unique identifier (\textbf{uid}), position (\textbf{x,y}) and health (\textbf{h}), constituting \textbf{struct\{uid, x, y, h\}}. An array is a collection elements of the same type, such as an integer or floating point number or a struct. An Array-of-Structures (AoS) can be constructed from structures, such as a group of units in a game. Alternatively, we can represent the data as a Structure-of-Arrays (SoA). For our unit example, this changes the data representation from \textbf{[struct\{uid, x, y, h\}, struct\{uid, x, y, h\}, ...]} to \textbf{struct\{[uid, uid, ...], [x, x, ...], [y, y, ...], [h, h, ...]\}}. Modern computers read from system memory in chunks, consequently accessing an individual unit's health will load the rest of the \textit{struct}. This wastes memory bandwidth on reading unused data, if, for example the user wanted to count the number of units with no health. Therefore, a SoA representation can reduce memory bandwidth usage by improving data locality; in this example only health data is loaded from system memory. Furthermore, contiguous grouping of the same property can make data more amenable to compression. The health property of a group of units with the same value is trivially compressible when contiguous.

\subsection{StarCraft II Datasets}

Many works have collected datasets of humans playing various video-games, intending to train deep learning models to either mimic or outperform humans \citen{smerdov2020collection, xenopoulos2022esta, lee2021feature}. \citet{SC2EGSet} generated a dataset from 55 StarCraft II tournaments, providing an API for the preprocessed data and a tool to analyze information such as player scores. However, their tool does not playback the game using the StarCraft II engine, and thus, lacks most of the game's state, which can only be recovered by replaying the game. Earlier works in this area, such as \citet{skillcraft_paper} who collected a dataset of 3,360 replays and processed the data into a CSV file. However, this dataset is relatively small and also lacks most of the game state information. \citet{synnaeve2012dataset} released a more detailed dataset with \~8,000 games; their dataset contains more game state information than previous works. However, this dataset is still relatively small compared to Blizzard Replay Packs, which contain over 900,000 games. \citet{wu2017msc} used the Blizzard Replay Packs and PySC2 \citen{vinyals2017starcraft} to create a numpy-backed dataset. They applied filtering techniques to ensure only higher quality replays would be included in their dataset. Our API is designed to replicate their filtering techniques using an auxiliary SQL database that can be used both to select replays for training and testing or summarize typical statistics on the replays as a whole, for example, the average game length or player Actions-Per-Minute (APM).

Some works do not release a dataset but rather the creation tool within the scope of other work. \citet{mathieu2023alphastar} processed a large amount of StarCraft II data to train \textit{AlphaStar-Unplugged}. However, their tool naively structures the data, producing data 10 times larger than our tool. Furthermore, they do not release the processed data, making the entry barrier very high for other researchers. \citet{leeExpert} released a tool to help generate synthetic data collected by running simulated agents against each other in simulated combat scenarios. However, this data does not contain any gameplay of human experts. In a different domain, \citet{kulinski2023starcraftimage} processed 60K replays to produce a dataset of minimaps for image processing, focusing on computer vision research. Our tool records a comprehensive view of the game state, including units, minimap, score, and economics data, akin to \citet{mathieu2023alphastar}.

\begin{table}
\centering
\caption{Existing Datasets in the Literature}
\begin{tabular}{lccccc}
\hline
Method                              & Replays & API & Mini Maps & Game Data & Data Source \\ \hline
\citen{SC2EGSet}                    & 17,930 & \ForestGreencheck & \Redcross & \ForestGreencheck & Tournaments \\
\citen{kulinski2023starcraftimage}  & 60,000   & \ForestGreencheck & \ForestGreencheck & \Redcross & Replay Packs \\
\citen{skillcraft_paper}            & 3,360    & \Redcross & \Redcross & \ForestGreencheck & Custom \\
\citen{wu2017msc}                   & 36,691   & \Redcross & \ForestGreencheck & \ForestGreencheck & Replay Packs \\
\citen{leeExpert}                   & 12,000   & \Redcross & \ForestGreencheck & \ForestGreencheck & Bots \\
\citen{synnaeve2012dataset}         & 7,649    & \Redcross\tablefootnote{API is no longer available.}    & \ForestGreencheck & \ForestGreencheck &  Tournaments         \\ \hline
\textbf{Ours} & \textbf{>900,000}   & \ForestGreencheck & \ForestGreencheck & \ForestGreencheck & \textbf{Tournament + Replay Pack} \\ \hline
\end{tabular}
\end{table}

\section{Dataset Overview} \label{sec:dataset}

We compiled two primary datasets using public game data from StarCraft II version 4.9.2\footnote{\url{https://github.com/Blizzard/s2client-proto/blob/master/samples/replay-api/download_replays.py}} and tournament data\footnote{\url{https://lotv.spawningtool.com/replaypacks/}}. The public dataset is significantly larger than the tournament dataset and comprises gameplay from a wide range of skill levels across the general player base. In contrast, the tournament dataset, drawn from various game versions, captures gameplay from skilled players operating at a notably higher level of expertise.

There are three stages of a dataset's life-cycle to analyze: creation, ownership and usage. Each stage plays a crucial role in optimizing costs, enhancing accessibility for users with budget constraints. A key factor in adaptability is ease of use; cumbersome or slow APIs can lead to user frustration, particularly in deep-learning pipelines where maintaining GPU throughput during model training is essential. Thus, streamlining these stages is paramount for reducing overall costs, improving accessibility, and ensuring a smoother experience for further research.

\subsection{Creation}

Our proposed serialization process, \textit{sc2-serializer}, transforms SC2 replay files from a sequential list of recorded player actions, into a series of game state observations usable for machine learning. This is achieved by re-simulating the game with the SC2 game engine, and querying the game via the Protobuf API to request state information. To determine the fastest possible conversion speed, observation requests are sent to the game, but the responses are discarded. This approach, referred to as \textit{game-speed}, represents the fastest possible conversion process. We use both \textit{game-speed} and \textit{AlphaStar-Unplugged} \citen{mathieu2023alphastar} as points of comparison for evaluation.

To overcome the task of recording millions of games, many conversion processes are run in parallel. The following benchmark considers this, and converts the same replay, repeated across multiple parallel processes on a 16-core Ryzen 5950X CPU with 64GB RAM. Table \ref{tab:parallel-bench} shows that \textit{sc2-serializer} operates closer to \textit{game-speed} than \textit{AlphaStar-Unplugged}, and can run more parallel processes before running Out-Of-Memory (OOM). The throughput plateau observed from $16\rightarrow24$ processes indicates that maximum throughput is achieved at the number of physical CPU cores. The chosen replay for this benchmark is a game of median duration. Longer replays proportionally require more memory, hence \textit{AlphaStar-Unplugged} will often trigger an OOM event with fewer than $16$ instances.

\begin{table}
    \centering
    \caption{Time to Process a Single Game Replay per Instance (sec) With N Instances}
    \begin{tabular}{lcccc}
        \hline
        Program &N=1 &N=8 &N=16 &N=24  \\
        \hline
        \textit{game-speed} (lower bound) &$43.8\pm0.3$ &$53.6\pm0.9$ &$83.5\pm2.0$  &$133.1\pm1.0$ \\
        \hline
        \textit{sc2-serializer} &$\mathbf{47.0\pm0.7}$ &$\mathbf{57.5\pm0.8}$ & $\mathbf{87.1\pm1.4}$ & $\mathbf{134.5\pm1.0}$ \\ 
        \textit{AlphaStar-Unplugged} &$61.8\pm1.6$ &$68.2\pm1.2$ &$87.2\pm1.2$ &OOM \\
        \hline
    \end{tabular}
    \label{tab:parallel-bench}
\end{table}

\begin{figure}
\centering
\includegraphics{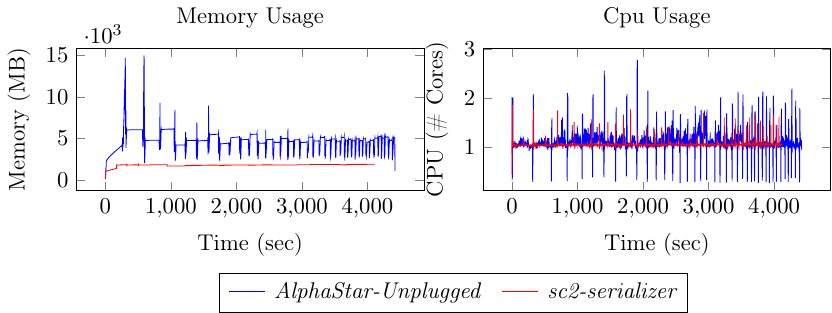}
\caption{Compute resources measured by querying Docker API when serializing 20 replays. \textit{AlphaStar-Unplugged} took 4217 seconds to complete whereas \textit{sc2-serializer} took 3828 seconds.}
\label{fig:compute-usage}
\end{figure}

To measure CPU and memory utilization, \textit{sc2-serializer} and \textit{AlphaStar-Unplugged} are run in a Docker container with resource utilization recorded using the Docker API, shown in Figure \ref{fig:compute-usage}. We converted 20 randomly sampled replays of various duration and observed that \textit{AlphaStar-Unplugged} exhibited significantly higher memory usage. Throughout the serialization process, both programs maintain a similar 1-CPU Core usage. However, \textit{sc2-serializer} completed in 3828 seconds, outperforming \textit{AlphaStar-Unplugged} which took 4217 seconds, representing a $\approx9\%$ time reduction. In general, \textit{AlphaStar-Unplugged} typically requires 8GB of memory per instance, although fluctuations can occur as in several of the larger memory spikes observed in Fig. \ref{fig:compute-usage}. Systems are typically equipped with about 4GB per physical CPU core, as observed from typical offerings from GCP \citen{gcp-vms}, AWS \citen{aws-vms} and within our lab. Hence, only half of the potential CPU throughput is attainable when using \textit{AlphaStar-Unplugged}. \textit{sc2-serializer} is consistently below this threshold (Fig. \ref{fig:compute-usage}), and therefore can run twice the number of conversion instances per machine. With our resources it took approximately one month to convert the 4.9.2 replay pack dataset. \textit{AlphaStar-Unplugged} would take more than double that time as half the number of instances could be run, and each instance $\approx11\%$ slower.

\subsection{Ownership} \label{sec:ownership}

\textbf{Storage:} Storage size is a crucial, yet often overlooked, aspect of the dataset life cycle. When a dataset fits within system memory, it can be kept in memory rather than read from disk each time, speeding up access. This speed up extends to the medium that the dataset is stored on -- solid-state storage offers significantly improved data accessing performance than hard-drive storage, at higher cost per capacity. Moreover, if the dataset is accessed over a network, smaller size reduces network bandwidth utilization. Therefore, it is important to reduce the on-disk size of a dataset when possible. We use ZLIB\citen{zlib} for compression when writing our replay data to disk. For the following experiments, we use 20 randomly sampled replays from the 492 dataset, biased towards longer duration replays.

\begin{figure}
    \centering
\includegraphics{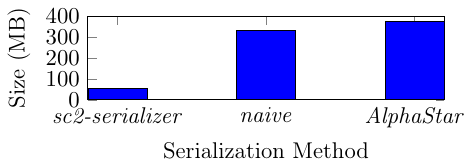}
    \caption{Serialization size of 20 randomly selected (biased towards longer duration) replays from StarCraft II. Note, \textit{sc2-serializer} includes a constant $16$MB index which becomes negligible.}
    \label{fig:size-comparison}
\end{figure}

The \textit{naive} approach to store StarCraft II replay data is to sequentially write observation data for each timestep. This results in a structure of $N_{timestep}\times N_{units}\times D_{unit}$: an array of $N_{timestep}$ timesteps, each containing a set of $N_{units}$ units with $D_{unit}$ attributes. If we were to naively write the observation data to disk in the same format it was gathered, we would only achieve marginal improvements in compression when compared to \textit{AlphaStar-Unplugged}. Therefore, we modify the data structures before serialization to a form that is more amenable to ZLIB, and observe a significant reduction in filesize (Fig. \ref{fig:size-comparison}).

Our objective is to increase the frequency of recurring patterns in the data by adjusting the data structure of in-game units. Using the original structure yields a filesize of $75.8$MB (Row 1, Table \ref{tab:unit_shape_compression}). Interestingly, a SoA transform at the timestep level reduces compressibility (Row 2, Table \ref{tab:unit_shape_compression}). However, a SoA transform that transposes $D_{unit}$ to the front leads to a more expected result; increasing compressibility (Row 3, Table \ref{tab:unit_shape_compression}). Further improvements can be achieved by grouping the second dimension by unit instance instead of time, reducing the file size to $21.6$MB (Row 4, Table. \ref{tab:unit_shape_compression}). We detail this process in Algorithm \ref{alg:instance-maj}, which can be applied to any data structure by changing $comp$ function to group by the desired property. The C++ implementation and the reverse transform is provided in the supplementary material (Appendix. \ref{code:instance-transform}). The improved compression ratio can be explained by a unit maintaining the same properties for extended periods, or indefinitely, such as the position of a building. The interleaving of units with static and dynamic data complicates the compression of the static data. However, when grouped by instance first, the repetition of static properties is maintained, thereby improving compressibility.

\begin{algorithm}
\caption{Transform to Instance-Major Structure-of-Arrays}\label{alg:instance-maj}
\begin{algorithmic}
\State \textbf{Input:} $sequenceData \gets$ Sequence of Observations, $comp \gets$ Binary Less Than Operator
\State \textbf{Output:} $(flatDataSoA, indices)$ Transformed Data and Original Index
\State $flatPairs \gets$ empty sequence of $(index,element)$ pairs
\For{$i \gets 0$ \textbf{to} $|sequenceData|-1$} \Comment{Flatten $sequenceData$ with timestep $i$}
    \For{$element$ \textbf{in} $sequenceData[i]$}
        \State \textbf{Append} $(i, element)$ \textbf{to} $flatPairs$
    \EndFor
\EndFor
\State stable\_sort($flatPairs$, $comp$) \Comment{Group instances by sorting by $InstanceId$}
\State $flatDataSoA \gets$ AoStoSoA(view\_elements($stepDataFlat$)) \Comment{Grouped instance data to SoA}
\State $indices \gets$ view\_indices($flatPairs$) \Comment{Copy time indices for later reconstruction}
\State \Return $(flatDataSoA, indices)$
\end{algorithmic}
\end{algorithm}

\begin{table}
    \centering
    \caption{Filesize of Unit Representation}
    \begin{tabular}{cc}
        \hline
        Order-Of-Dimensions                             & Size [MB] \\
        \hline
        $N_{timestep}\times N_{units}\times D_{unit}$   & 75.8      \\
        $N_{timestep}\times D_{unit}\times N_{units}$   & 97.3      \\
        $D_{unit}\times N_{timestep}\times N_{units}$ (\textit{time-major})     & 53.6      \\
        $D_{unit}\times N_{units}\times N_{timestep}$ (\textit{instance-major}) & \textbf{21.8} \\
        \hline
    \end{tabular}
    \label{tab:unit_shape_compression}
\end{table}

The same batch of $20$ replays converted by \textit{AlphaStar-Unplugged} yields $383.2$MB. In contrast, \textit{sc2-serializer} yields $39.5$MB, a factor of $\approx10$ reduction. Furthermore, multiple replays contained in a file reduces pressure on the OS file-system to index many smaller files. The final size of our 4.9.2 replay pack dataset is $1$TB, using \textit{AlphaStar-Unplugged} would balloon this to $\approx10$TB.

\begin{figure}
\centering
\includegraphics{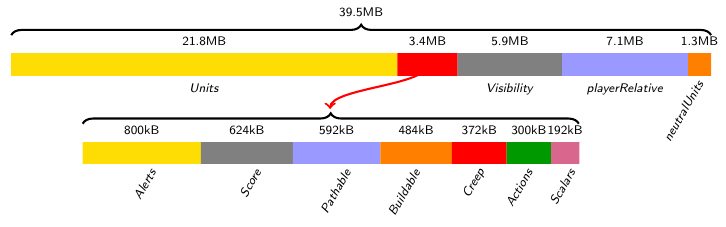}
\caption{Filesize contribution breakdown of 20 randomly selected replays after serialization.
}
\label{fig:data-contrib}
\end{figure}

Our efforts to reduce the overall size of the dataset was guided by analysing the contribution the components which make up the dataset's size. This lead us to identify that the unit data had a significant contribution of $75.8$MB. The final breakdown of the data contribution of each source is shown in Figure \ref{fig:data-contrib}, depicting the \textit{Units} and \textit{neutralUnits} after the instance-major transformation at $21.8$MB and $1.3$MB respectively. Minimap feature layers shown in Figure \ref{fig:data-contrib} include \textit{visibility}, \textit{playerRelative}, \textit{Alerts}, \textit{Pathable}, \textit{Buildable} and \textit{Creep}. Minimaps that represent binary values, such as \textit{Creep}, are packed in 8-bit characters for serialization, as boolean arrays are not a native datatype. The remaining components, \textit{Scalars}, \textit{Score} and \textit{Actions} are structurally simple and have a low contribution to the filesize.

\textbf{Deserialization}: Reading data from disk and transforming it into a desired format quickly and efficiently is a critical characteristic for dataset usability. GPUs training deep-learning based models must be efficiently fed with data to train on. In the context of StarCraft II replay files and machine learning tasks, efficient deserialization is essential to access game state observations stored on disk. The original data that our StarCraft II dataset is generated from is an order of magnitude smaller, as it contains only a list of actions made by the players. However, to obtain game state observations for machine learning, the game must be causally re-simulated. This is many orders of magnitude greater in time and compute resources than reading from our serialized format.

In our methodology, a crucial additional step involves recovering the unit data in timestep-major form, for convenient access to unit information at each timestep. This would ordinarily result in a more expensive dataloader, but similar deserialization time to the timestep-major approach is observed because the file-reading stage of deserialization is faster for the smaller files that our instance-major approach produces. To benchmark our approach against \textit{AlphaStar-Unplugged}, which uses TFRecord files, we analyzed deserialization of a median duration replay file. Results are reported with mean and standard deviation over 100 trials. Our format demonstrates a deserialization speed similar to \textit{AlphaStar-Unplugged}, along with substantially reduced file size. However, we note that this experiment was conducted on local NVMe storage, a lower bandwidth, higher latency medium such as HDD or network storage would favor our smaller format. Hence, our approach maintains efficiency in data retrieval, while also reducing storage requirements, a compelling advantage.

\begin{table}
    \centering
    \caption{Time taken to de-serialize a replay from each player's perspective}
    \begin{tabular}{lll}
    \hline
    Algorithm           & Player 1 (ms)             & Player 2 (ms)            \\
    \hline
    sc2-serializer      & \textbf{44.04$\pm$1.99}   & \textbf{53.16$\pm$2.56}  \\
    AlphaStar-Unplugged & 54.76$\pm$9.36            & 54.26$\pm$3.14           \\
    \hline
    \end{tabular}
    \label{tab:deserialize}
\end{table}

\subsection{Usage}

Python bindings to C++ code are used to efficiently manipulate data for machine learning. The simplest and smallest observation data structures are designed to appear first in the serialized replays. Therefore, if a user wants to learn based on the game score or economy, only this information is deserialized. Minimaps are next in order, then finally the unit data, the largest contributor in size and complexity in the dataset. We take advantage of this in our own research (Sec. \ref{section:training}), where we primarily use scalar and minimap data. Additionally, a metadata header enables quick extraction into a complimentary SQL database. This enables filtering of replays during data loading, offering a pragmatic means to identify and skip low-quality games. For instance, criteria such as game duration or player activity levels can be effectively utilized as filters, aiding in curating a higher-quality dataset. We refer the reader to the supplementary material (Appendix \ref{sec:features}) for a more comprehensive discussion of features included in our framework.

\section{Training Experiments} \label{section:training}

\subsection{Outcome Prediction}

Outcome prediction considers the binary classification problem of which player will win~\citen{SC2EGSet,kulinski2023starcraftimage,vinyals2017starcraft}. Similar to \citet{SC2EGSet}, we fit a unique XGBoost \citen{XGBoost} model at regular time intervals to predict the outcome of a game from the player's Point-of-View (POV). We also train a set of deep learning models that use a CNN encoder for minimap data and MLP encoder for scalar data. The latent feature vectors from each encoder are concatenated and decoded by another MLP to produce the final outcome prediction. The weights of the deep learning models are shared across time points. However, scalar inputs are pre-processed with a BatchNorm layer per time point for normalization, since data such as minerals collected increases over time. We also include a `Global POV' setting where the model has access to both players' POV and predicts which one will win.\footnote{Experiment source code available at \url{https://github.com/5had3z/sc2-experiments}}

To compare against existing methods, we use prediction results at the 15 minute mark of a game. Other works utilize different inputs, datasets and models, therefore, results are not directly comparable. Results presented in Table \ref{tab:outcome} highlight the effectiveness of our larger dataset, compared to other works.

\begin{table}
\centering
\caption{Game Outcome Prediction Results}
\label{tab:outcome}
\begin{tabular}{llcccccc}
\hline
\multirow{2}{*}{Work}      & \multirow{2}{*}{Algorithm} & Global    & \multicolumn{2}{c}{Input} & \multicolumn{2}{c}{Accuracy at 15 min.} \\ \cline{4-7}
                           &                            & POV       & Scalar           & Minimap                    & 4.9.2 & Tourn.  \\
\hline
\multirow{3}{*}{\cite{SC2EGSet}}  & SVM                 & \Redcross & \ForestGreencheck & \Redcross         & -     & 67.4\%  \\ 
                           & XGBoost                    & \Redcross & \ForestGreencheck & \Redcross         & -     & 67.9\% \\
                           & Linear Reg.                & \Redcross & \ForestGreencheck & \Redcross         & -     & 69.5\% \\
\hline
\cite{vinyals2017starcraft}       & CNN                 & \Redcross & \ForestGreencheck & \ForestGreencheck & 65\%  & - \\ 
\hline
\cite{kulinski2023starcraftimage} & CNN                 & \Redcross & \Redcross         & \ForestGreencheck & 59.4\%\tablefootnote{Mean accuracy over all time} & - \\
\hline
\multirow{4}{*}{Ours}      & XGBoost                    & \Redcross         & \ForestGreencheck & \Redcross         & 74.0\% & 67.3\% \\
                           & CNN                        & \Redcross & \Redcross         & \ForestGreencheck & 72.3\% & 69.3\% \\
                           & MLP                        & \Redcross & \ForestGreencheck & \Redcross         & 71.6\% & 69.3\% \\
                           & CNN+MLP                    & \Redcross & \ForestGreencheck & \ForestGreencheck & 75.9\% & 69.4\% \\ 
                           & CNN+MLP                    & \ForestGreencheck & \ForestGreencheck & \ForestGreencheck & \textbf{82.0}\% & \textbf{80.6\%} \\ 

\hline          
\end{tabular}
\end{table}

\begin{figure}
    \centering
\includegraphics{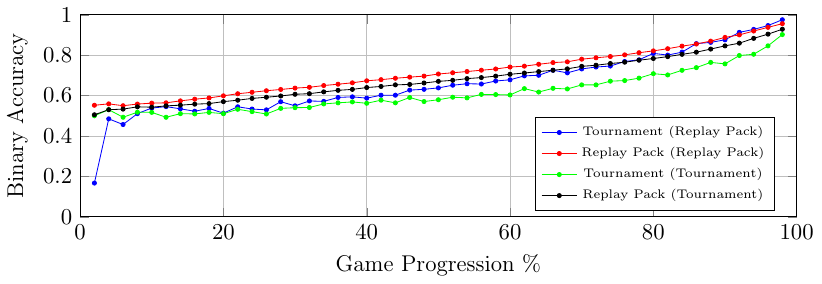}
    \caption{Evaluation using the CNN+MLP Global POV model on both in-distribution and out-of-distribution data. Legend is formatted as: \textit{Training Set (Testing Set)}.}
    \label{fig:gamelength}
\end{figure}

Results align with \citet{SC2EGSet} when employing the tournament dataset. However, using the extensive dataset of Replay pack demonstrates a noticeable performance boost. However, a limitation of evaluation at a fixed 15 minute time point is that it heavily depends on the number of games concluding at a given time. This dependency arises because predicting game outcomes becomes considerably easier as they near completion. To address this concern, following \citet{sanchez-predict-games}, we represent results as a percentage of game length (Fig. \ref{fig:gamelength}). This approach offers a more insightful perspective on our model's performance. Our analysis shows that training on larger datasets (Replay packs) and evaluating on tournaments outperforms training and evaluating solely on tournaments. This emphasizes the significance of large datasets and efficient data generation tools.


\subsection{Minimap Forecasting}

The in-game minimap of StarCraft II is a birds-eye-view of the entire map that depicts information such as map topology, player visibility, and the relative alliance between visible units. Events are also highlighted to draw the player's attention, such as when they are being attacked. Forecasting the future minimap demonstrates the capability to model motion dynamics of the units. \citet{kulinski2023starcraftimage} explores this task with their ``hyper-spectral'' image dataset by predicting the next ``hyper-spectral'' image conditioned on the previous. We explore a more direct variation, predicting the future sequence of ego player unit occupancy on the minimap. $27$ second clips of the tournament dataset with an $\approx3$ second sample period are used\footnote{The periodicty of the data varies since the replay observations are only sampled on player actions}. The challenge is to predict $3$ steps into the future, conditioned on the previous $6$. The observation space includes the height map, occupancy of the ego player, and the occupancy of the enemy player (under the visibility rules of the game).

A variety of architectures are tested, including Siamese-ResNets decoded with a 3D-convolution (3D-Conv), a Spatio-Temporal Vision Transformer (ViT), and a UNet\citen{nnUnet} (frames are concatenated along the channel dimension). Predicting pixel occupancy is performed either directly or as a residual from the previous frame. For a residual prediction, tanh is applied to the output logits, rather than sigmoid, then added to the last observation and clamped between $0$ and $1$. The model learns to emit $-1$ to transition pixels to unoccupied, $0$ for no change, and $1$ for newly occupied. Pixel-wise binary cross entropy is used as the loss function and optimized with AdamW\citen{loshchilov2018decoupled}. To prioritize modeling the occupancy of units in motion, a pixel-wise weighting factor is applied in two ways. One applies weighting where pixels differ from the `Last' observation. Another applies weighting if there has been `Motion' in that pixel over the sequence.

Area Under the Curve (AUC) and Soft Intersection-over-Union (Soft IoU) are used to measure performance. A trivial baseline is included that repeats the last observed minimap as a prediction. A very high Soft IoU score can be achieved with the aforementioned baseline, hence, masked Soft IoU variants are included to better measure forecasting quality for units in motion. Here, a `Static' mask removes pixels that are occupied for the entire sequence (static buildings) and a `Last' mask removes pixels that are equal to the last observation, thereby measuring the change between images. An example of a minimap occupancy sequence and the evaluation masks is included in the supplementary material.

\begin{table}
\centering
\caption{Predicting the Future Minimap of the Ego-Player's Units}
\label{tab:minimap-perf}
\resizebox{0.99\textwidth}{!}{%
{\renewcommand{\arraystretch}{1.2} 
\begin{tabular}{l c ccc ccc ccc ccc}
\hline
\multirow{2}{*}{Algorithm} & Pixel Loss 
    &\multicolumn{3}{c}{Soft IoU}
    &\multicolumn{3}{c}{Soft IoU - Static}
    &\multicolumn{3}{c}{Soft IoU - Last}
    &\multicolumn{3}{c}{AUC} \\ 
\cline{3-14} & Weighting
    & +3s & +6s & +9s
    & +3s & +6s & +9s
    & +3s & +6s & +9s
    & +3s & +6s & +9s \\
\hline
Propagate Last & - 
    & \textbf{0.772} & \textbf{0.757} & \textbf{0.737} 
    & \textbf{0.348} & \textbf{0.319} & \textbf{0.282} 
    & - & - & - 
    & 0.763 & 0.747 & 0.726 \\
ViT & Last 
    & 0.382 & 0.384 & 0.385 
    & 0.171 & 0.176 & 0.180 
    & \textbf{0.328} & \textbf{0.340} & \textbf{0.346} 
    & 0.742 & 0.743 & 0.743 \\ 
3D-Conv & Last 
    & 0.533 & 0.517 & 0.371 
    & 0.185 & 0.169 & 0.123 
    & 0.184 & \underline{0.152} & \underline{0.174} 
    & 0.886 & 0.870 & 0.746  \\  
UNet Residual & Last 
    & 0.691 & 0.667 & 0.646
    & 0.251 & 0.222 & 0.198
    & \underline{0.267} & \underline{0.152} & 0.157
    & \underline{0.948} & \textbf{0.935} & \textbf{0.923} \\ 
UNet Residual & Motion 
    & 0.727 & 0.702 & 0.677
    & 0.284 & 0.248 & 0.220
    & 0.169 & 0.125 & 0.130
    & 0.946 & 0.929 & 0.917  \\ 
UNet Residual & - 
    & 0.732 & 0.706 & 0.685
    & 0.238 & 0.242 & 0.217
    & 0.169 & 0.124 & 0.128
    & 0.947 & 0.930 & 0.917  \\ 
UNet & Motion 
    & \underline{0.738} &\underline{0.711} & \underline{0.689}
    & 0.302 & 0.261 & 0.234
    & 0.182 & 0.137 & 0.140
    & \textbf{0.949} & \underline{0.933} & \underline{0.921}  \\ 
\hline 
\end{tabular}}}
\end{table}

\begin{figure}
    \centering
    \begin{tabular}{ccc}
        \subfloat[Last Frame]{\includegraphics[width=0.3\textwidth]{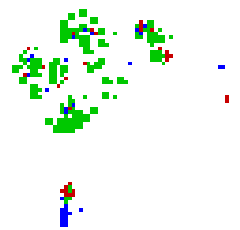}} &
        \subfloat[UNet]{\includegraphics[width=0.3\textwidth]{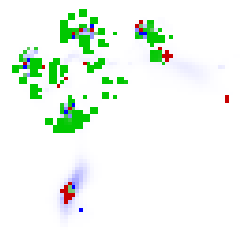}} &
        \subfloat[ViT]{\includegraphics[width=0.3\textwidth]{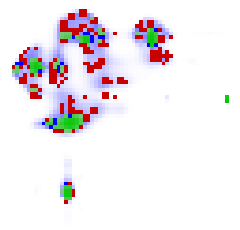}} \\
    \end{tabular}
    \caption{Forecasted Ego-agent Unit Occupancy on the minimap $3$ sec into the future. True-positive threshold is $P>0.5$ and shown in \textcolor{green}{green}, false negative is shown \textcolor{red}{red}, and false-positive as \textcolor{blue}{blue}. }
    \label{fig:minimap-pred}
\end{figure}

The UNet outperforms the baseline in AUC, a measure of how correctly ordered predictions are, i.e. the highest scoring pixels are correct. Even, when masking static pixels, players tend to move units in groups from one spot to another, becoming idle once they reach their destination. Outperforming the naive baseline in Soft-IoU is a challenge, as the model must be well calibrated in its estimation. Hence, poor Soft-IoU performance is a result of poor uncertainty calibration, indicating that a more precise learning of output logit scaling is required. The UNet performed better and is cheaper to train than the other models. This is attributable to its higher spatial resolution, resulting in sharper prediction than the ViT (Fig. \ref{fig:minimap-pred}). Figure \ref{fig:minimap-pred} does show more accurate prediction from the ViT for the moving army, but its lower resolution produces a softer output resulting in worse overall performance.

\section{Limitations and Ethical Considerations} \label{sec:discussion}

A limitation of this dataset, and others \citen{mathieu2023alphastar}, is that replay data is not randomly accessible due to compression. Therefore, even if only intending to sample a few time steps, the entire replay must be de-serialized. This is a problem when training small models on sub-sequences of data, the cost of loading the data is greater than the training iteration. In our outcome-prediction experiment, we pre-process our data into simple numpy (.npz) files which contain samples for training. A potential solution could be to serialize in chunks to enable skipping over and deserializing of individual chunks. However, the utility of this feature is difficult to gauge since the optimal chunk size will vary on a case-by-case basis. nvCOMP\footnote{\url{https://developer.nvidia.com/nvcomp}} could also be used to accelerate deserialization, however the uncompressed replay would then consume GPU Memory otherwise needed for training. These issues don't arise when training larger models with limited throughput, or if using the entire replay.

Akin to many other machine learning libraries, we chose to write our library in C++ due to its high-performance. We have endeavored to design the library as customizeable for application in scenarios other than StarCraft II through template meta-programming. Those who wish to re-apply the framework on a novel dataset will require familiarity with C++. However, this could act as a barrier to entry since this programming language is less commonly used in the machine learning community. Furthermore, the generated data is unable to be read by any tool other than our framework. StarCraft II library users do not need to delve into C++ since all APIs have python bindings (with type hinting).

The dataset provided is primarily suited for the training of AI agents and testing of algorithms in gaming settings. SC2 is a popular game for research in many fields due to its accessible API that enables observing and interacting with the game. Although there are no immediate ramifications or concerns arising from our contribution that we are aware of, the approaches and techniques developed using the serializer and dataset could be used in a range of downstream applications ranging from potentially controversial military simulations to digital twin dataset curation in manufacturing.


\section{Conclusion}

We introduce a StarCraft II replay serialization tool that reduces the dataset size by a factor of $\approx10$ and more than halves the creation time for systems with a typical memory-to-compute ratio. This improves accessibility for researchers with constrained storage and compute resources. These efficiencies also reduce the overall carbon footprint of working with StarCraft II data. We demonstrate this greater amount of data improves the accuracy of outcome forecasting models by $11\%$, and enables motion forecasting models to predict $9$ seconds into the future with $0.923$ AUC. Our large and detailed dataset enables further research in complex real-time strategy games. The same optimizations can be applied to scenarios where data consists of objects with a range of properties that change at different timescales, for example a set of warehousing robots with static unchanging data (robot type), slowly changing data (payload), and rapidly changing data (position) (see Appendix \ref{sec:features-dataset} for an example). By rearranging this data for rapid compression and fast disk read/writes, we reduce the cost of ownership.

\impact{Reducing the costs associated with complex dataset creation, ownership and usage has a broad range of benefits for the research community and society as a whole. A lower cost of ownership improves accessibility for institutions and individual researchers who are resource constrained, empowering them to work on complex multi-agent behavioural tasks such as StarCraft II. As mentioned in our introduction, there are domains with related structures, namely intralogistics modeling, which has been understudied and lacks public datasets. Our framework would aid in creating such a dataset which is amenable for training a deep learning based model on. Furthermore, a reduction of dataset costs contributes to lower storage and energy costs, reducing the carbon footprint of deep learning.
 
However, more efficient storage also reduces the cost of ownership and improves the accessibility of mass data collection for institutions with malicious intentions such as oppressive governments. This can enable large scale data collection and analysis of the unaware public, violating their privacy and potentially lead to targeted oppression. However this data is usually in tabular form, where formats such as parquet are already used to store and access efficiently.}

\acks{This work was supported by an Australian Government Research Training Program (RTP) Scholarship.}

\bibliography{refs}

\begin{thebibliography}{32}
\providecommand{\natexlab}[1]{#1}
\providecommand{\url}[1]{\texttt{#1}}
\expandafter\ifx\csname urlstyle\endcsname\relax
  \providecommand{\doi}[1]{doi: #1}\else
  \providecommand{\doi}{doi: \begingroup \urlstyle{rm}\Url}\fi

\bibitem[Amazon()]{aws-vms}
Amazon.
\newblock Amazon ec2 instance types.
\newblock \url{https://aws.amazon.com/ec2/instance-types}, 2024.
\newblock Accessed: 2024-05-22.

\bibitem[Bia{\l}ecki et~al.(2023)Bia{\l}ecki, Jakubowska, Dobrowolski, Bia{\l}ecki, Krupi{\'n}ski, Szczap, Bia{\l}ecki, and Gajewski]{SC2EGSet}
Andrzej Bia{\l}ecki, Natalia Jakubowska, Pawe{\l} Dobrowolski, Piotr Bia{\l}ecki, Leszek Krupi{\'n}ski, Andrzej Szczap, Robert Bia{\l}ecki, and Jan Gajewski.
\newblock {SC2EGSet}: {StarCraft} {II} esport replay and game-state dataset.
\newblock \emph{Sci Data}, 10\penalty0 (1):\penalty0 600, September 2023.

\bibitem[Cao et~al.(2023)Cao, Kreis, Fidler, Sharp, and Yin]{cao2023texfusion}
Tianshi Cao, Karsten Kreis, Sanja Fidler, Nicholas Sharp, and Kangxue Yin.
\newblock Texfusion: Synthesizing 3d textures with text-guided image diffusion models.
\newblock In \emph{Proceedings of the IEEE/CVF International Conference on Computer Vision}, pages 4169--4181, 2023.

\bibitem[Chen and Guestrin(2016)]{XGBoost}
Tianqi Chen and Carlos Guestrin.
\newblock Xgboost: A scalable tree boosting system.
\newblock In \emph{Proceedings of the 22nd ACM SIGKDD International Conference on Knowledge Discovery and Data Mining}, KDD '16, page 785–794, New York, NY, USA, 2016. Association for Computing Machinery.
\newblock ISBN 9781450342322.
\newblock \doi{10.1145/2939672.2939785}.
\newblock URL \url{https://doi.org/10.1145/2939672.2939785}.

\bibitem[Deng et~al.(2009)Deng, Dong, Socher, Li, Li, and Fei-Fei]{ImageNET}
Jia Deng, Wei Dong, Richard Socher, Li-Jia Li, Kai Li, and Li~Fei-Fei.
\newblock Imagenet: A large-scale hierarchical image database.
\newblock In \emph{2009 IEEE Conference on Computer Vision and Pattern Recognition}, pages 248--255, 2009.
\newblock \doi{10.1109/CVPR.2009.5206848}.

\bibitem[Drissi~Elbouzidi et~al.(2023)Drissi~Elbouzidi, Ait El~Cadi, Pellerin, Lamouri, Tobon~Valencia, and B{\'e}langer]{drissi2023role}
Adnane Drissi~Elbouzidi, Abdessamad Ait El~Cadi, Robert Pellerin, Samir Lamouri, Estefania Tobon~Valencia, and Marie-Jane B{\'e}langer.
\newblock The role of ai in warehouse digital twins: literature review.
\newblock \emph{Applied sciences}, 13\penalty0 (11):\penalty0 6746, 2023.

\bibitem[ESL()]{katowice-prize}
ESL.
\newblock Your ultimate guide to iem sc2 katowice 2024.
\newblock \url{https://pro.eslgaming.com/tour/2024/02/your-ultimate-guide-to-iem-sc2-katowice-2024}, 2024.
\newblock Accessed: 2024-05-22.

\bibitem[Futrega et~al.(2022)Futrega, Milesi, Marcinkiewicz, and Ribalta]{nnUnet}
Micha{\l} Futrega, Alexandre Milesi, Micha{\l} Marcinkiewicz, and Pablo Ribalta.
\newblock Optimized u-net for brain tumor segmentation.
\newblock In Alessandro Crimi and Spyridon Bakas, editors, \emph{Brainlesion: Glioma, Multiple Sclerosis, Stroke and Traumatic Brain Injuries}, pages 15--29, Cham, 2022. Springer International Publishing.
\newblock ISBN 978-3-031-09002-8.

\bibitem[Gailly and Adler(2004)]{zlib}
Jean-loup Gailly and Mark Adler.
\newblock zlib compression library.
\newblock 12 2004.

\bibitem[Galli et~al.(2011)Galli, Loiacono, Cardamone, and Lanzi]{fpsCheating}
Luca Galli, Daniele Loiacono, Luigi Cardamone, and Pier~Luca Lanzi.
\newblock A cheating detection framework for unreal tournament iii: A machine learning approach.
\newblock In \emph{2011 IEEE Conference on Computational Intelligence and Games (CIG'11)}, pages 266--272. IEEE, 2011.

\bibitem[Gallotta et~al.(2024)Gallotta, Todd, Zammit, Earle, Liapis, Togelius, and Yannakakis]{videoGameLLM}
Roberto Gallotta, Graham Todd, Marvin Zammit, Sam Earle, Antonios Liapis, Julian Togelius, and Georgios~N Yannakakis.
\newblock Large language models and games: A survey and roadmap.
\newblock \emph{arXiv preprint arXiv:2402.18659}, 2024.

\bibitem[Google()]{gcp-vms}
Google.
\newblock General-purpose machine family for compute engine.
\newblock \url{https://cloud.google.com/compute/docs/general-purpose-machines}, 2024.
\newblock Accessed: 2024-05-22.

\bibitem[Harmer et~al.(2018)Harmer, Gisslén, del Val, Holst, Bergdahl, Olsson, Sjöö, and Nordin]{battlefieldRL}
Jack Harmer, Linus Gisslén, Jorge del Val, Henrik Holst, Joakim Bergdahl, Tom Olsson, Kristoffer Sjöö, and Magnus Nordin.
\newblock Imitation learning with concurrent actions in 3d games.
\newblock In \emph{2018 IEEE Conference on Computational Intelligence and Games (CIG)}, pages 1--8, 2018.
\newblock \doi{10.1109/CIG.2018.8490398}.

\bibitem[Hoffmann et~al.(2022)Hoffmann, Borgeaud, Mensch, Buchatskaya, Cai, Rutherford, de~Las~Casas, Hendricks, Welbl, Clark, Hennigan, Noland, Millican, van~den Driessche, Damoc, Guy, Osindero, Simonyan, Elsen, Vinyals, Rae, and Sifre]{NEURIPS2022_c1e2faff}
Jordan Hoffmann, Sebastian Borgeaud, Arthur Mensch, Elena Buchatskaya, Trevor Cai, Eliza Rutherford, Diego de~Las~Casas, Lisa~Anne Hendricks, Johannes Welbl, Aidan Clark, Thomas Hennigan, Eric Noland, Katherine Millican, George van~den Driessche, Bogdan Damoc, Aurelia Guy, Simon Osindero, Kar\'{e}n Simonyan, Erich Elsen, Oriol Vinyals, Jack Rae, and Laurent Sifre.
\newblock An empirical analysis of compute-optimal large language model training.
\newblock In S.~Koyejo, S.~Mohamed, A.~Agarwal, D.~Belgrave, K.~Cho, and A.~Oh, editors, \emph{Advances in Neural Information Processing Systems}, volume~35, pages 30016--30030. Curran Associates, Inc., 2022.
\newblock URL \url{https://proceedings.neurips.cc/paper_files/paper/2022/file/c1e2faff6f588870935f114ebe04a3e5-Paper-Conference.pdf}.

\bibitem[Huffman(1952)]{huffman1952method}
David~A Huffman.
\newblock A method for the construction of minimum-redundancy codes.
\newblock \emph{Proceedings of the IRE}, 40\penalty0 (9):\penalty0 1098--1101, 1952.

\bibitem[Kulinski et~al.(2023)Kulinski, Waytowich, Hare, and Inouye]{kulinski2023starcraftimage}
Sean Kulinski, Nicholas~R Waytowich, James~Z Hare, and David~I Inouye.
\newblock Starcraftimage: A dataset for prototyping spatial reasoning methods for multi-agent environments.
\newblock In \emph{Proceedings of the IEEE/CVF Conference on Computer Vision and Pattern Recognition}, pages 22004--22013, 2023.

\bibitem[Lee and Ahn(2021)]{lee2021feature}
Chan~Min Lee and Chang~Wook Ahn.
\newblock Feature extraction for starcraft ii league prediction.
\newblock \emph{Electronics}, 10\penalty0 (8):\penalty0 909, 2021.

\bibitem[Lee et~al.(2021)Lee, Kim, and Ahn]{leeExpert}
Donghyeon Lee, Man-Je Kim, and Chang~Wook Ahn.
\newblock Predicting combat outcomes and optimizing armies in starcraft ii by deep learning.
\newblock \emph{Expert Systems with Applications}, 185:\penalty0 115592, 2021.
\newblock ISSN 0957-4174.
\newblock \doi{https://doi.org/10.1016/j.eswa.2021.115592}.
\newblock URL \url{https://www.sciencedirect.com/science/article/pii/S0957417421009921}.

\bibitem[Loshchilov and Hutter(2019)]{loshchilov2018decoupled}
Ilya Loshchilov and Frank Hutter.
\newblock Decoupled weight decay regularization.
\newblock In \emph{International Conference on Learning Representations}, 2019.
\newblock URL \url{https://openreview.net/forum?id=Bkg6RiCqY7}.

\bibitem[M{\"a}rtens et~al.(2015)M{\"a}rtens, Shen, Iosup, and Kuipers]{toxicGameplay}
Marcus M{\"a}rtens, Siqi Shen, Alexandru Iosup, and Fernando Kuipers.
\newblock Toxicity detection in multiplayer online games.
\newblock In \emph{2015 International Workshop on Network and Systems Support for Games (NetGames)}, pages 1--6. IEEE, 2015.

\bibitem[Mathieu et~al.(2023)Mathieu, Ozair, Srinivasan, Gulcehre, Zhang, Jiang, Paine, Powell, Żołna, Schrittwieser, Choi, Georgiev, Toyama, Huang, Ring, Babuschkin, Ewalds, Bordbar, Henderson, Colmenarejo, van~den Oord, Czarnecki, de~Freitas, and Vinyals]{mathieu2023alphastar}
Michaël Mathieu, Sherjil Ozair, Srivatsan Srinivasan, Caglar Gulcehre, Shangtong Zhang, Ray Jiang, Tom~Le Paine, Richard Powell, Konrad Żołna, Julian Schrittwieser, David Choi, Petko Georgiev, Daniel Toyama, Aja Huang, Roman Ring, Igor Babuschkin, Timo Ewalds, Mahyar Bordbar, Sarah Henderson, Sergio~Gómez Colmenarejo, Aäron van~den Oord, Wojciech~Marian Czarnecki, Nando de~Freitas, and Oriol Vinyals.
\newblock Alphastar unplugged: Large-scale offline reinforcement learning, 2023.

\bibitem[Qu et~al.(2022)Qu, Li, and Qian]{JetClass}
Huilin Qu, Congqiao Li, and Sitian Qian.
\newblock {Particle Transformer} for jet tagging.
\newblock In \emph{{Proceedings of the 39th International Conference on Machine Learning}}, pages 18281--18292, 2022.

\bibitem[Smerdov et~al.(2020)Smerdov, Zhou, Lukowicz, and Somov]{smerdov2020collection}
Anton Smerdov, Bo~Zhou, Paul Lukowicz, and Andrey Somov.
\newblock Collection and validation of psychophysiological data from professional and amateur players: A multimodal esports dataset.
\newblock \emph{arXiv preprint arXiv:2011.00958}, 2020.

\bibitem[Statistica(2024)]{gameTAM}
Statistica.
\newblock \url{https://www.statista.com/outlook/dmo/digital-media/video-games/worldwide}, 2024.
\newblock Accessed: 2024-05-07.

\bibitem[Synnaeve and Bessiere(2012)]{synnaeve2012dataset}
Gabriel Synnaeve and Pierre Bessiere.
\newblock A dataset for starcraft ai \& an example of armies clustering, 2012.

\bibitem[Sánchez-Ruiz(2015)]{sanchez-predict-games}
Antonio Sánchez-Ruiz.
\newblock Predicting the winner in two player starcraft games.
\newblock 1394:\penalty0 24--35, 01 2015.

\bibitem[Thompson et~al.(2013)Thompson, Blair, Chen, and Henrey]{skillcraft_paper}
Joseph~J. Thompson, Mark~R. Blair, Lihan Chen, and Andrew~J. Henrey.
\newblock Video game telemetry as a critical tool in the study of complex skill learning.
\newblock \emph{PLOS ONE}, 8\penalty0 (9):\penalty0 1--12, 09 2013.
\newblock \doi{10.1371/journal.pone.0075129}.
\newblock URL \url{https://doi.org/10.1371/journal.pone.0075129}.

\bibitem[Vinyals et~al.(2017)Vinyals, Ewalds, Bartunov, Georgiev, Vezhnevets, Yeo, Makhzani, Küttler, Agapiou, Schrittwieser, Quan, Gaffney, Petersen, Simonyan, Schaul, van Hasselt, Silver, Lillicrap, Calderone, Keet, Brunasso, Lawrence, Ekermo, Repp, and Tsing]{vinyals2017starcraft}
Oriol Vinyals, Timo Ewalds, Sergey Bartunov, Petko Georgiev, Alexander~Sasha Vezhnevets, Michelle Yeo, Alireza Makhzani, Heinrich Küttler, John Agapiou, Julian Schrittwieser, John Quan, Stephen Gaffney, Stig Petersen, Karen Simonyan, Tom Schaul, Hado van Hasselt, David Silver, Timothy Lillicrap, Kevin Calderone, Paul Keet, Anthony Brunasso, David Lawrence, Anders Ekermo, Jacob Repp, and Rodney Tsing.
\newblock Starcraft ii: A new challenge for reinforcement learning, 2017.

\bibitem[Vinyals et~al.(2019)Vinyals, Babuschkin, Czarnecki, Mathieu, Dudzik, Chung, Choi, Powell, Ewalds, Georgiev, Oh, Horgan, Kroiss, Danihelka, Huang, Sifre, Cai, Agapiou, Jaderberg, Vezhnevets, Leblond, Pohlen, Dalibard, Budden, Sulsky, Molloy, Paine, Gulcehre, Wang, Pfaff, Wu, Ring, Yogatama, W{\"u}nsch, McKinney, Smith, Schaul, Lillicrap, Kavukcuoglu, Hassabis, Apps, and Silver]{Vinyals2019}
Oriol Vinyals, Igor Babuschkin, Wojciech~M. Czarnecki, Micha{\"e}l Mathieu, Andrew Dudzik, Junyoung Chung, David~H. Choi, Richard Powell, Timo Ewalds, Petko Georgiev, Junhyuk Oh, Dan Horgan, Manuel Kroiss, Ivo Danihelka, Aja Huang, Laurent Sifre, Trevor Cai, John~P. Agapiou, Max Jaderberg, Alexander~S. Vezhnevets, R{\'e}mi Leblond, Tobias Pohlen, Valentin Dalibard, David Budden, Yury Sulsky, James Molloy, Tom~L. Paine, Caglar Gulcehre, Ziyu Wang, Tobias Pfaff, Yuhuai Wu, Roman Ring, Dani Yogatama, Dario W{\"u}nsch, Katrina McKinney, Oliver Smith, Tom Schaul, Timothy Lillicrap, Koray Kavukcuoglu, Demis Hassabis, Chris Apps, and David Silver.
\newblock Grandmaster level in starcraft ii using multi-agent reinforcement learning.
\newblock \emph{Nature}, 575\penalty0 (7782):\penalty0 350--354, Nov 2019.
\newblock ISSN 1476-4687.
\newblock \doi{10.1038/s41586-019-1724-z}.
\newblock URL \url{https://doi.org/10.1038/s41586-019-1724-z}.

\bibitem[Wu et~al.(2017)Wu, Zhang, and Huang]{wu2017msc}
Huikai Wu, Junge Zhang, and Kaiqi Huang.
\newblock Msc: A dataset for macro-management in starcraft ii.
\newblock \emph{arXiv preprint arXiv:1710.03131}, 2017.

\bibitem[Xenopoulos and Silva(2022)]{xenopoulos2022esta}
Peter Xenopoulos and Claudio Silva.
\newblock Esta: An esports trajectory and action dataset, 2022.

\bibitem[Ziv and Lempel(1977)]{ziv1977universal}
Jacob Ziv and Abraham Lempel.
\newblock A universal algorithm for sequential data compression.
\newblock \emph{IEEE Transactions on information theory}, 23\penalty0 (3):\penalty0 337--343, 1977.

\end{thebibliography}

\appendix
\section{}

\subsection{Dataset Statistics}

\begin{figure}[htbp]
    \centering
    \begin{subfigure}{0.3\textwidth}
        \centering
        \resizebox{\linewidth}{!}{
\includegraphics{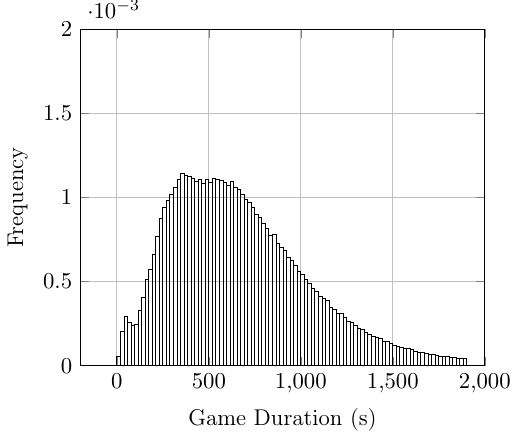}
            }
    \end{subfigure}
    \begin{subfigure}{0.3\textwidth}
        \centering
        \includegraphics[width=1\linewidth]
        {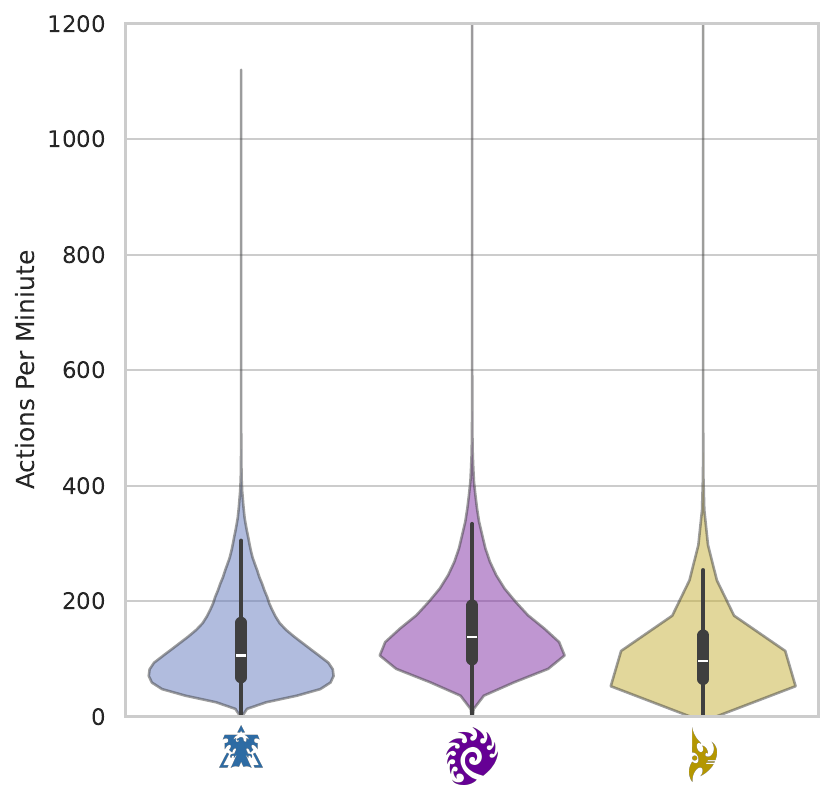}
    \end{subfigure}
    \begin{subfigure}{0.3\textwidth}
        \centering
        \includegraphics[width=1\linewidth]{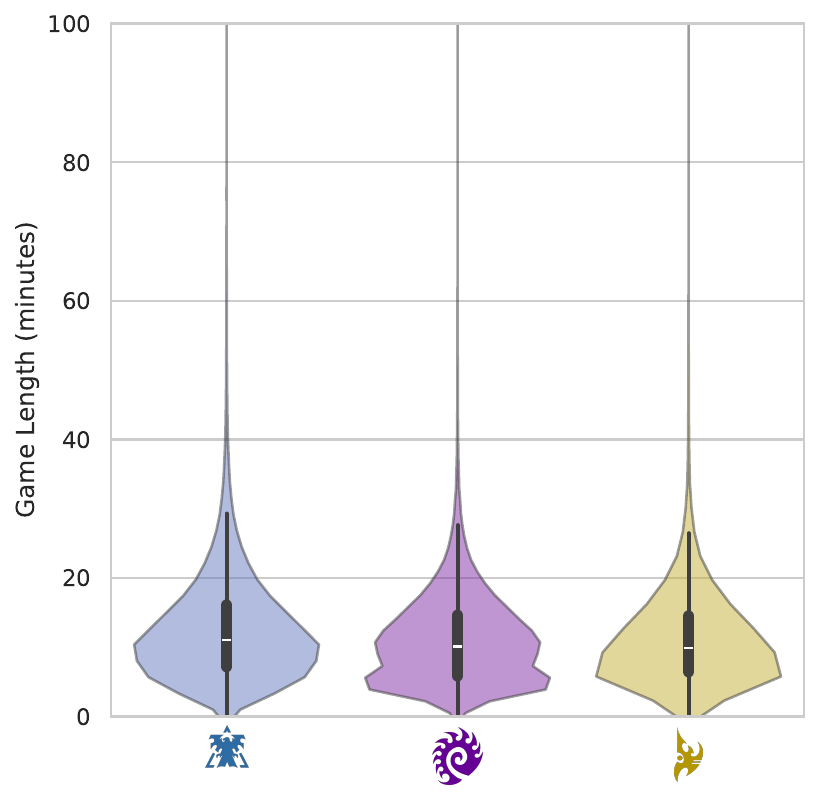}
    \end{subfigure}
    \begin{subfigure}{0.3\textwidth}
        \centering
        \resizebox{\linewidth}{!}{

\includegraphics{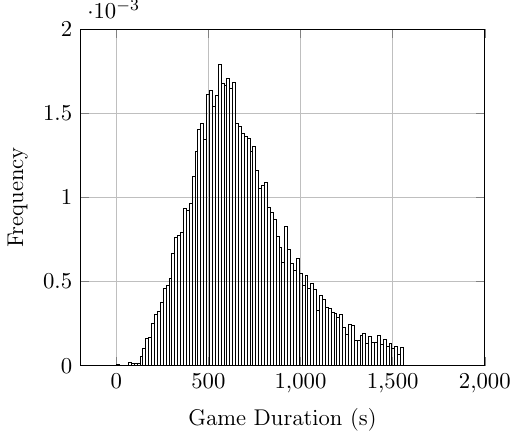}
            }
    \end{subfigure}
    \begin{subfigure}{0.3\textwidth}
        \centering
        \includegraphics[width=1\linewidth]{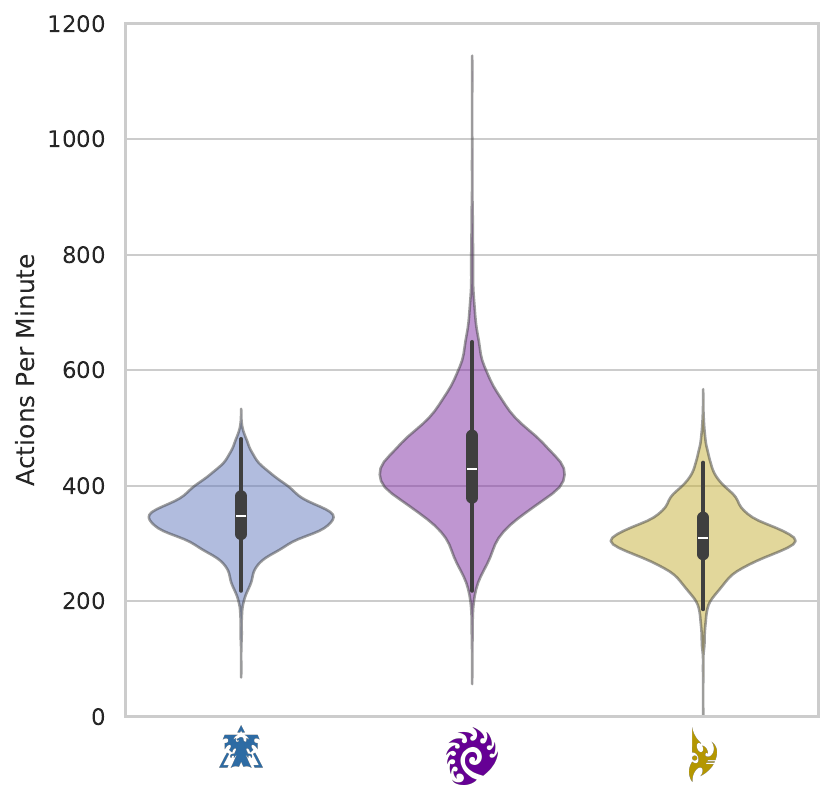}
    \end{subfigure}
    \begin{subfigure}{0.3\textwidth}
        \centering
        \includegraphics[width=1\linewidth]{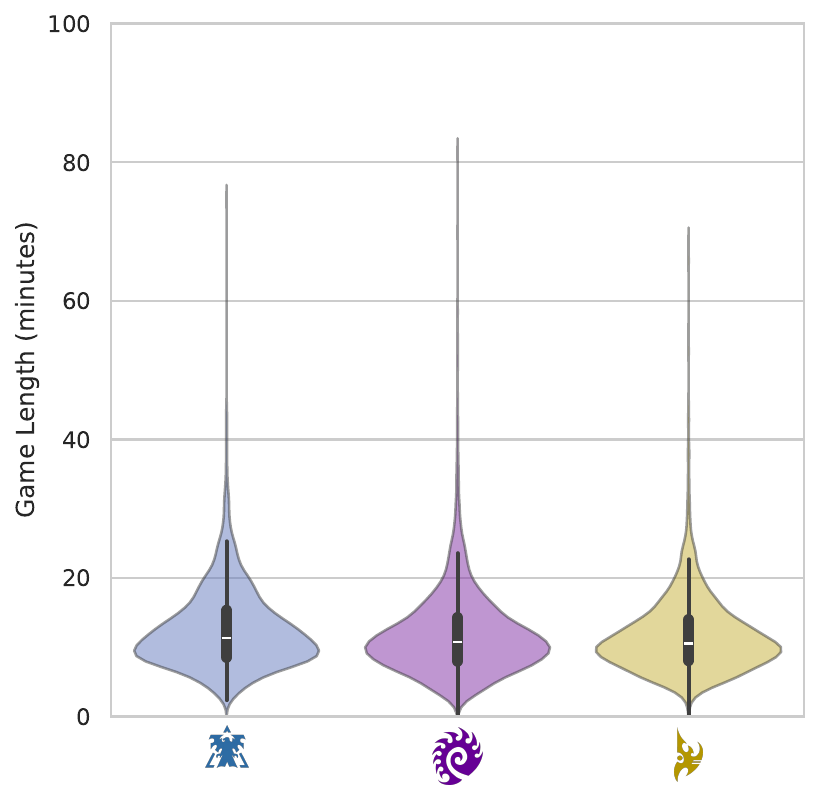}
    \end{subfigure}
    \caption{Statistics such as Game Duration (Left), APM (Center) and Faction-Based Time-to-Win (Right) for both public matches from the 4.9.2 Replay Pack (Top) and gathered Tournament Replays (Bottom).}
    \label{fig:data-stats}
\end{figure}

To gain some insights into the data we perform statistical comparisons between the public and professional tournament data (Fig. \ref{fig:data-stats}). From these figures, we can glean high level knowledge, for example that the Actions-Per-Minute (APM) of professional players is typically higher than the general public, and Zerg players have a higher APM than other factions. Interestingly, the duration for both public and professional games are similar in distribution with $695\pm430$s and $718\pm355$s mean and standard deviation respectively. Another interesting phenomenon can be observed in the time to win for each faction depicted in the right column. There is a notable bi-modal distribution for Zerg faction in the public data, which often wins in the early game. This is expected as the Zerg faction is renowned for being aggressive in early game. However, professional players expect this strength and are able to counter effectively, hence the winning time distribution is more even between the factions in the tournament data. 

\subsection{Framework Features} \label{sec:features}

\subsubsection{Dataset Serialization} \label{sec:features-dataset}

Our serialization and dataloading library includes a variety of features to improve usability and customizability. We intend our serialization framework to be modified and applied to other games or simulations. Ergonomic features are implemented to improve user experience for typical operations such as transformations and filtering when dataloading for machine learning.

Several code documentation frameworks are utilized to enhance navigability and ease-of-use. Python code contains docstrings and type hints. C++ structures and functions exported to Python include automatically generated Python type hints to enable intellisense when using the library in Python. C++ code is documented with Doxygen. All the documentation is collated and rendered with sphinx, centralizing both the C++ and Python APIs, enabling users to gain an overall understanding how the code fits together. This helps users to identify where changes are needed to adapt to their scenario. Furthermore, we provide examples on how to apply the library to novel scenarios. Documentation is hosted via Github Pages\footnote{\url{https://5had3z.github.io/sc2-serializer/index.html}}.

We ensured the flexibility of the library as the proposed serialisation strategy is likely to benefit a range of settings where data follows an SC2 like setting. For example, consider a warehouse robot with the following data collected... \textbf{struct\{robot\_id, robot\_type, x\_pos, y\_pos, payload\_id, payload\_weight, destination\_id, battery\_charge, need\_maintenance, need\_assistance\}}. This data structure contains some highly dynamic data, \textbf{x\_pos, y\_pos}, moderately dynamic data \textbf{battery\_charge, payload\_id, need\_assistance} and static data \textbf{robot\_id, robot\_type}. We provide an example of applying our library in a novel scenario in the code repository.

Our framework is modularized, with separate components for replay observation, serialization and data structures. This separation enables customization in each area without changes to fix inter-dependencies. This modularity is achieved through C++ template meta-programming. For example, the replay database class, is templated on a replay data structure, which implements an interface for reading and writing to a standard input-output stream. The replay data structure can be modified, or a new variant created, with no changes to the replay database. We provide an example of using this library in a simple smart-factory-like scenario which records observations of a set of pedestrians and robots.

A significant portion of the data manipulation code is written in C++ for best performance. We leverage a basic reflection library, pfr\footnote{\url{https://github.com/boostorg/pfr}}, to write generic algorithms that iterate over data structures. This enables the creation or modification of data structures without changing or adding new manipulation algorithms. This is used in both the serialization and vectorization algorithms. To further increase performance, compile-time calculations are used for control flow, rather than at runtime, reducing pressure on the instruction cache and branch predictor. As an example, we tag structures with properties such as \textit{hasUnits} or \textit{hasMinimaps} which is checked with \textit{if constexpr()} to enable the unit or minimap parsing code path. This trivially enables different permutations of the parser to be compiled dependent on the data structure's tags. 

\subsubsection{StarCraft II}

The conversion tool from~\citet{mathieu2023alphastar} samples replay observations only when the player makes an action. We provide several sampling modes: ``on every step'', ``on every action'', ``every $n$ steps'', and a combination of on ''every $n$ steps and every action''. Results for conversion time and serialization size for \textit{sc2-serializer} are reported with ``on every action'' for a fair comparison with \textit{AlphaStar-Unplugged}.

Economic resources, minerals and vespene gas, play a critical role in StarCraft II. An artifact of the game's engine is that the UID of units change when they transition between being inside and outside of the player's vision. Hence, the UID of economic resources change each time they enter and exit the player's vision as units move around the map. This is a subtlety that may go unnoticed by algorithms that rely on UIDs. We remedy this by checking for a positional match with existing instances, then associate the new UID to the original UID. Furthermore, the quantity remaining of economic resources are set to zero when outside of the player's vision. We instead recall the last known quantity observed by the player, and update this quantity whenever there is a new observation. For resources outside of the player's vision at the beginning of the match, we set all resources to the expected default.

Another feature of our framework, absent from others, includes calculating a one-hot encoding of active upgrades for a player. Offline, we query the Blizzard API for all the possible player actions. The response from the API includes a unique action identifier, and some text description. We register actions as an upgrade if it resembles an upgrade from the wiki\footnote{\url{https://starcraft.fandom.com/wiki/StarCraft_Wiki}, changes between game versions and discrepancies in naming schemes made this a non-trivial task}. For each faction and version of the game, a one-hot encoding of upgrades is constructed. At runtime, the time step when the upgrade is active is recorded. To get the current state of active upgrades, a simple comparison $gameStep>upgradeFinishedTime$ is used to create a one-hot encoding of the active upgrades.

\subsection{Instance Transform Code} \label{code:instance-transform}

\begin{minted}[fontsize=\footnotesize,]{c}   
/**
 * @brief First version of property-major data sorting, includes index
 * with each element as a separate vector, used to
 * recover the original time-major format.
 *
 * @tparam SoA structure type of flattened data
 * @tparam Comp comparison function type, the input datatype to sort
 * is std::pair<std::uint32_t,SoA::struct_type>
 * @param stepData data to rearrange
 * @param comp Comparison function used to sort the data
 * @return FlattenedData<SoA> data sorted by property with accompanying
 * original time index
 */
template<IsSoAType SoA, typename Comp>
[[nodiscard]] auto flattenAndSortData(
    const std::vector<std::vector<typename SoA::struct_type>> &stepData,
    Comp &&comp) noexcept -> FlattenedData<SoA>
{
    using element_t = SoA::struct_type;
    using index_element_t = std::pair<std::uint32_t, element_t>;
    std::vector<index_element_t> stepDataFlat;
    for (std::size_t idx = 0; idx < stepData.size(); ++idx) {
        std::ranges::transform(
            stepData[idx], std::back_inserter(stepDataFlat),
            [=](element_t u) { return std::make_pair(idx, u); });
    }
    // Significantly better compressibility when sorted by unit
    // (and implicitly time)
    std::ranges::stable_sort(stepDataFlat, comp);
    // Create flattened SoA
    auto dataSoA = AoStoSoA<SoA>(std::views::values(stepDataFlat));
    // Create accompanying step indices for reconstruction
    std::vector<uint32_t> indices(stepDataFlat.size());
    std::ranges::copy(std::views::keys(stepDataFlat), indices.begin());
    return { dataSoA, indices };
}

/**
 * @brief Transform instance-major unit data back to time-major
 * @tparam UnitSoAT
 * @param flattenedUnits instance-major data to transform
 * @return Unit data grouped by time
 */
template<IsSoAType SoA>
[[nodiscard]] auto recoverFlattenedSortedData(
    const FlattenedData<SoA> &stepDataFlat) noexcept 
    -> std::vector<std::vector<typename SoA::struct_type>>
{
    // Create outer dimension with the maximum game step index
    const std::size_t maxStepIdx = std::ranges::max(stepDataFlat.indices);
    std::vector<std::vector<typename SoA::struct_type>> stepData(
        maxStepIdx + 1ull);
    // Copy units to correct timestep
    const auto &indices = stepDataFlat.indices;
    for (std::size_t idx = 0; idx < indices.size(); ++idx) {
        stepData[indices[idx]].emplace_back(stepDataFlat.data[idx]);
    }
    return stepData;
}
\end{minted} 

\subsection{Minimap Evaluation Protocol}

To evaluate the performance of minimap forecasting, our Soft-IoU calculation includes a pixel-wise mask $m$ to isolate the prediction accuracy of pixels that change between the \textbf{Last} and next frame, or to remove the contribution of \textbf{Static} units,

\begin{equation}
Soft IoU=\frac{\sum_{N_{pixel}}o\hat{o}m}{\sum_{N_{pixel}}(o+\hat{o}-o\hat{o})m}.
\label{eq:soft_iou}
\end{equation}

\begin{figure}
    \centering
    \begin{tabular}{ccc}
        \subfloat[Input (t=-2)]{\includegraphics[width=0.3\textwidth]{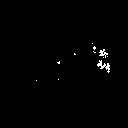}} &
        \subfloat[Input (t=-1)]{\includegraphics[width=0.3\textwidth]{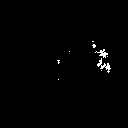}} &
        \subfloat[Input (t=0)]{\includegraphics[width=0.3\textwidth]{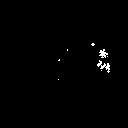}} \\
        \subfloat[Target (t=1)]{\includegraphics[width=0.3\textwidth]{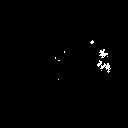}} &
        \subfloat[Static Mask]{\includegraphics[width=0.3\textwidth]{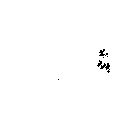}} &
        \subfloat[Last Mask]{\includegraphics[width=0.3\textwidth]{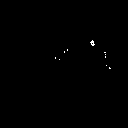}} \\
    \end{tabular}
    \caption{ The top row of binary masks depicts the observed occupancy sequence. Based on these, the model is to predict the next occupancy mask (Bottom Left). To isolate the prediction accuracy of units in motion, \textbf{Static} units are masked (Bottom Center). Furthermore, to isolate accuracy of predicting the change between two images, pixels are masked if they are equal between the \textbf{Last} and next frame (Bottom Right). }
    \label{fig:minimap-seq}
\end{figure}

Where $\hat{o}$ is the predicted occupancy of a pixel, and $o$ is the ground truth occupancy. Figure \ref{fig:minimap-seq} depicts these masks and an input sequence for the model.

\end{document}